\documentclass[letterpaper, 10 pt, conference]{ieeeconf}
\IEEEoverridecommandlockouts

\let\labelindent\relax
\let\labelindent\relax
\usepackage{graphicx} \graphicspath{ {figures/} }
\usepackage{amsmath,amssymb,mathabx,amsbsy,mathtools,etoolbox}
\usepackage[utf8]{inputenc} 
\usepackage[T1]{fontenc}    

\usepackage{algorithmic}
\usepackage[ruled,linesnumbered]{algorithm2e}
\usepackage{acronym}
\usepackage{enumitem}
\usepackage[breaklinks,linkcolor=black]{hyperref}
\usepackage{balance}
\usepackage{xspace}
\usepackage{setspace}
\usepackage[skip=3pt,font=small]{subcaption}
\usepackage[skip=3pt,font=small]{caption}
\usepackage[dvipsnames,svgnames,x11names]{xcolor}
\usepackage[capitalise,nameinlink]{cleveref}
\usepackage{booktabs,tabularx,colortbl,multirow,multicol,array,makecell}
\usepackage{pifont}
\usepackage{cite}
\usepackage{nicematrix}
\usepackage{comment}

\makeatletter
\DeclareRobustCommand\onedot{\futurelet\@let@token\@onedot}
\def\@onedot{\ifx\@let@token.\else.\null\fi\xspace}

\def\ie{\emph{i.e}\onedot}

\makeatother

\makeatletter
\def\BState{\State\hskip-\ALG@thistlm}
\makeatother

\makeatletter
\renewcommand{\paragraph}{%
  \@startsection{paragraph}{4}%
  {\z@}{0ex \@plus 0ex \@minus 0ex}{-1em}%
  {\hskip\parindent\normalfont\normalsize\bfseries}%
}
\makeatother

\crefname{algorithm}{Alg.}{Algs.}
\Crefname{algocf}{Algorithm}{Algorithms}
\crefname{section}{Sec.}{Secs.}
\Crefname{section}{Section}{Sections}
\crefname{table}{Tab.}{Tabs.}
\Crefname{table}{Table}{Tables}
\crefname{figure}{Fig.}{Fig.}
\Crefname{figure}{Figure}{Figure}

\definecolor{gblue}{HTML}{4285F4}
\definecolor{gred}{HTML}{DB4437}
\definecolor{ggreen}{HTML}{0F9D58}

\definecolor{mygray}{gray}{.92}

\acrodef{qp}[QP]{Quadratic Programming}
\acrodef{fd}[FD]{Force Decomposition}
\acrodef{ros}[ROS]{Robot Operating System}
\acrodef{uav}[UAV]{Unmanned Aerial Vehicle}
\acrodef{dof}[DoF]{Degree-of-freedom}
\acrodef{com}[CoM]{Center-of-Mass}
\acrodef{ilc}[ILC]{Iterative Learning Control}
\acrodef{siso}[SISO]{single-input-single-output} 
\acrodef{zpetc}[ZPETC]{zero-phase-error tracking controller}
\acrodef{mimo}[MIMO]{multi-input-multi-output}

\makeatletter
\renewcommand*{\@opargbegintheorem}[3]{\trivlist
  \item[\hskip \labelsep{\bfseries #1\ #2}] \textbf{(#3)}\ \itshape}
\makeatother

\let\oldnl\nl
\newcommand{\nonl}{\renewcommand{\nl}{\let\nl\oldnl}}%
\makeatother

\title{\LARGE \bf Probabilistic Visibility-Aware Trajectory Planning for Target Tracking in Cluttered Environments}

\author{Han Gao$^1$, Pengying Wu$^1$, Yao Su$^2$, Kangjie Zhou$^1$, Ji Ma$^1$, Hangxin Liu$^2$ and Chang Liu$^{1,\dagger}$
\thanks{$^\ast$This work was sponsored by Beijing Nova Program (20220484056) and the National Natural Science Foundation of China (62203018).}
\thanks{$\dagger$ Corresponding author. $^{1}$ Department of Advanced Manufacturing and Robotics, College of Engineering, Peking University. Emails: \{hangaocoe, littlefive, kangjiezhou, maji, changliucoe\}@pku.edu.cn. $^{2}$ National Key Laboratory of General Artificial Intelligence, Beijing Institute for General Artificial Intelligence (BIGAI). Emails: \{suyao, liuhx\}@bigai.ai.}}

\begin{document}

\maketitle
\begin{abstract}
Target tracking has numerous significant civilian and military applications, and maintaining the visibility of the target plays a vital role in ensuring the success of the tracking task. 
Existing visibility-aware planners primarily focus on keeping the target within the limited field of view of an onboard sensor and avoiding obstacle occlusion. However, the negative impact of system uncertainty is often neglected, rendering the planners delicate to uncertainties in practice. 
To bridge the gap, this work proposes a real-time, non-myopic trajectory planner for visibility-aware and safe target tracking in the presence of system uncertainty. For more accurate target motion prediction, we introduce the concept of belief-space probability of detection (BPOD) to measure the predictive visibility of the target under stochastic robot and target states. An Extended Kalman Filter variant incorporating BPOD is developed to predict target belief state under uncertain visibility within the planning horizon. To reach real-time trajectory planning, we propose a computationally efficient algorithm to uniformly calculate both BPOD and the chance-constrained collision risk by utilizing linearized signed distance function (SDF), and then design a two-stage strategy for lightweight calculation of SDF in sequential convex programming. Extensive simulation results with benchmark comparisons show the capacity of the proposed approach to robustly maintain the visibility of the target under high system uncertainty. The practicality of the proposed trajectory planner is validated by real-world experiments. 
\end{abstract}

\setstretch{0.98}
\section{Introduction}\label{sec1}
\begin{figure}[!t]
  \centering
  \includegraphics[width=\linewidth]{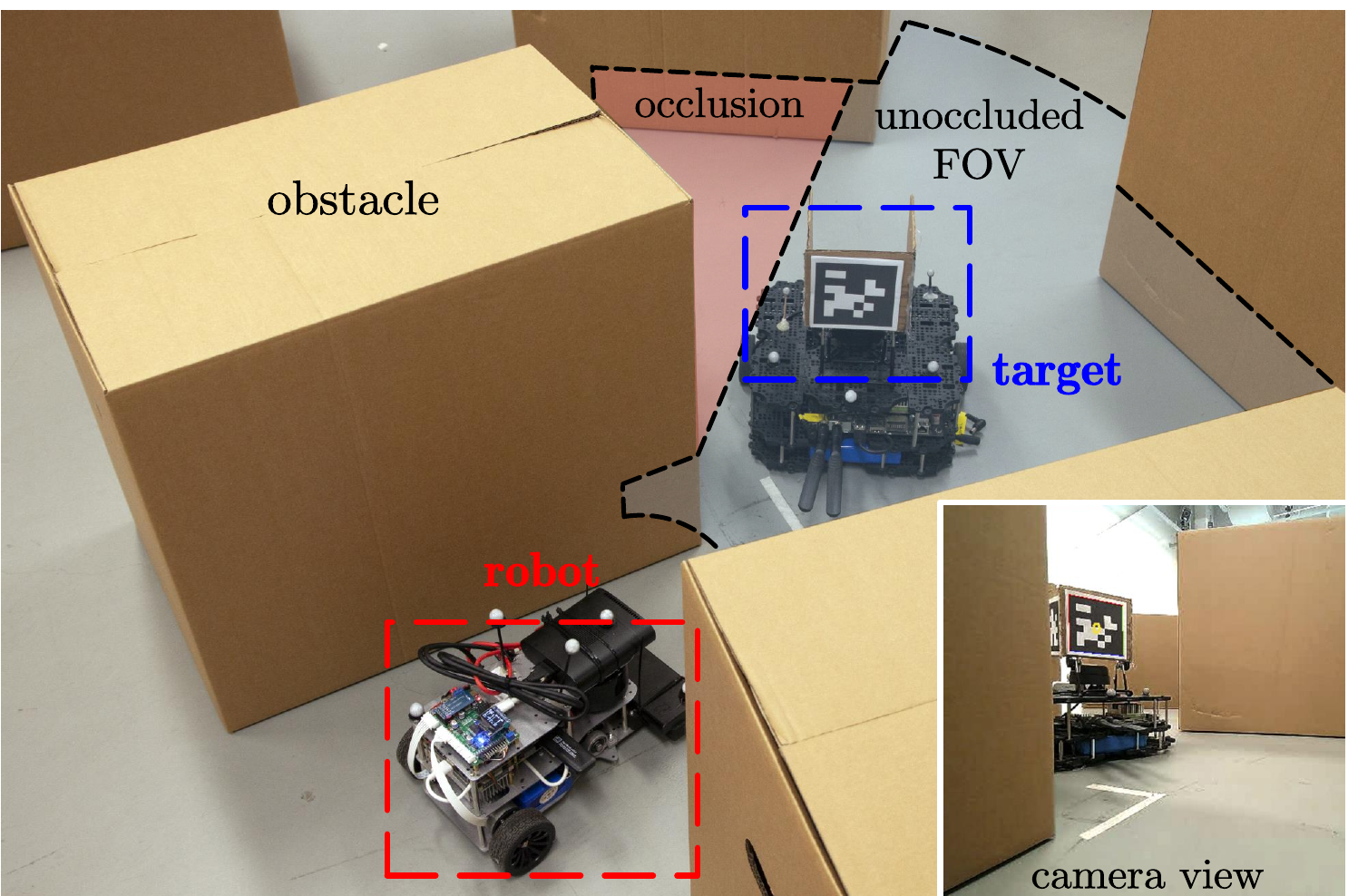}
  \caption{\textbf{Illustration of visibility-aware tracking for a moving target.} The red region represents the part of FOV being occluded, while the light blue region means the unoccluded part. Note that the occluded regions overlapping with obstacles are not visualized.}
 \label{head illus} 
\end{figure}

Target tracking using an autonomous vehicle has garnered widespread utilization in various important applications, such as vehicle tracking~\cite{punyavathi2022vehicle}, cinematography~\cite{xu2022interactive}, and underwater monitoring~\cite{xanthidis2021aquavis}. In recent years, visibility-aware motion planning has emerged as a key focus of target-tracking research, where the robot is tasked with generating trajectories to track a mobile target while ensuring continuous visibility, as illustrated in \cref{head illus}. 

Since first characterized by LaValle et al.~\cite{lavalle1997motion}, various motion planning methodologies have been introduced to tackle the visibility-aware tracking task. 
Some previous works formulate the target tracking problem as a visibility-based pursuer-evader game~\cite{zou2018optimal,lozano2022visibility} and focus on proposing the winning strategy for the pursuers to keep uninterrupted visibility of a target. Nevertheless, the adversarial behaviors of the target greatly complicate robot motion planning and lead to high computational costs \cite{lozano2022visibility,lozano2022surveillance}, thus hindering the application to real-time non-adversarial tracking tasks in realistic, complex environments.

Trajectory optimization, on the contrary, remains a mainstream methodology for real-time visibility-aware planning. This approach formulates trajectory planning as an optimization problem~\cite{su2022object}, seeking to maintain the target's visibility by adjusting robot control inputs~\cite{liu2017model, Li2022SMA} or predictive trajectory parameters~\cite{wang2021visibility}. Our work adopts this methodology due to its scalability and computational efficiency. 

Previous works on visibility-aware trajectory planning mainly focus on incorporating explicit considerations for the limited field of view (FOV) and obstacle occlusion, as they will directly impair the visibility of a target. To avoid potential target loss due to a limited FOV, previous works employ geometric costs, such as distance~\cite{xanthidis2021aquavis,wang2021visibility} and bearing angle~\cite{penin2018vision,wang2021visibility} costs between the target and the sensor, to maintain the target within the FOV. To handle possible occlusion by obstacles, previous works propose to apply distance cost between the line of sight (LOS) and obstacles to avoid occlusion~\cite{penin2018vision,masnavi2022visibility}. However, most of the abovementioned research overlooks the detrimental effects of system uncertainties 
arising from imperfect system models, process noise, and measurement noise, which may significantly degrade state estimation and trajectory planning in visibility-aware tracking tasks. 

Belief space planning (BSP)~\cite{patil2014gaussian} provides a systematic approach for handling system uncertainty, and is widely used in target tracking tasks with stochasticity~\cite{liu2017model,Li2022SMA}. Specifically, BSP formulates a stochastic optimization problem to generate visibility maintenance trajectories, and the key step lies in the evaluation of predictive target visibility. 
A commonly employed method for predicting visibility is to evaluate the probability of detection (POD) by integrating predictive target probability density function (PDF) over the anticipated unoccluded FOV area~\cite{yu2014cooperative,zhang2019multiple}. 
However, calculating this integration can be either computationally burdensome when conducted over continuous state spaces~\cite{zhang2019multiple} or inaccurate due to discretization error when transformed into the sum of probability mass over a grid detectable region~\cite{yu2014cooperative}.
Another method is directly determining target visibility based on the mean positions of the predicted target and the planned FOV~\cite{liu2017model, Li2022SMA}. Despite its computational efficiency, such maximum-a-posteriori (MAP) estimation only considers mean positions, neglecting predictive target uncertainty during robot trajectory planning.

This work proposes a model predictive control (MPC)-based non-myopic trajectory planner for mobile target tracking in cluttered environments. The proposed framework systematically considers limited FOV, and obstacle occlusion under a BSP framework to address state uncertainty. The main contributions can be summarized as follows:
\begin{itemize}
    \item We propose the concept of \textit{belief-space probability of detection} (BPOD) that depends on both robot and target belief states to function as a measure of predicted visibility. We then develop an Extended Kalman Filter (EKF) variant that incorporates BPOD into target state prediction to take into account stochastic visibility in the MPC predictive horizon, which overcomes the deficiency of MAP estimation. 
    \item We present a unified representation for both the BPOD and the collision risk as the \textit{probabilities of stochastic SDF satisfaction} (PoSSDF) via the use of signed distance functions (SDFs), and develop a paradigm to efficiently calculate the PoSSDFs.
    \item We propose a real-time trajectory planner for visibility-aware target tracking in cluttered environments based on sequential convex programming (SCP). In particular, we propose a two-stage strategy for calculating SDF in SCP to accelerate the planner, reaching a computational speed of $10Hz$. 
    Simulations with benchmark comparisons and real-world experiments demonstrate that our method enables the robot to track a moving target in cluttered environments at high success rates and visible rates, even under high system uncertainty.
\end{itemize}

The remainder of the article is organized as follows. \cref{sec2} formulates the target tracking problem. \cref{sec3} defines BPOD and proposes a variant of EKF covariance update formulation. \cref{sec4} approximates BPOD and collision risk by SDF linearization and proposes an SCP framework with a two-stage strategy for online trajectory planning. Simulations and real-world experiments are presented in \cref{sec5,sec6}, respectively. \cref{sec7} presents conclusions and future works.

\section{Formulation of Target Tracking}\label{sec2}
This work considers a mobile target tracking problem in a 2D environment with cluttered obstacles, as shown in \cref{head illus}. The robot knows its current state but relies on a noisy sensor with limited FOV to detect a stochastically moving target. Therefore, the target state remains partially observable, necessitating estimation by the robot. Besides, the motion model of the robot and the target are both stochastic, adding to the difficulty of target tracking. The goal of the robot is to plan collision-free trajectories to maintain continual detection of the target. 
\subsection{Motion Models and Obstacle Modeling}
The robot and target dynamics are defined as follows~\cite{zhou2024aspire}:
\begin{IEEEeqnarray}{rClCl}\small
\mathbf{z}^r_{k+1}&=&\mathbf{f}^r(\mathbf{z}^r_{k},\mathbf{u}^r_k)&+\mathbf{\mathbf{w}}^r,\ \   \mathbf{w}^r\sim \mathcal{N}(\mathbf{0},\mathbf{R}^r),\label{rbt motion model}\\
\mathbf{z}^t_{k+1}&=&\mathbf{f}^t(\mathbf{z}^t_{k},\mathbf{u}^t_k)&+\mathbf{w}^t,\ \  \mathbf{w}^t \sim \mathcal{N}(\mathbf{0},\mathbf{R}^t),\label{target motion model}
\end{IEEEeqnarray}
where $\mathbf{z}$ and $\mathbf{u}$ denote the state and control input, respectively. The superscripts $r$ and $t$ denote robot and target, respectively, and the subscript $k$ means time step. The motion function $\mathbf{f}^r$ and $\mathbf{f}^t$ will be specified in \cref{sec5}. The process noise $\mathbf{w}^r$ and $\mathbf{w}^t$ follow Gaussian distributions with zero means and covariance matrix $\mathbf{R}^r$ and $\mathbf{R}^t$, respectively.

 This work considers a known continuous map with convex obstacles represented as point sets $\mathcal{O}_i\subset \mathbb{R}^2, i=1,2,\cdots, N_o$, where $N_o$ denotes the number of obstacles. Note that the proposed approach can also address nonconvex obstacles by dividing them into multiple convex ones.

\subsection{Sensor Model}

The robot's sensor has a limited FOV $\mathcal{V}_k\subseteq \mathbb{R}^2$, which will be specified in \cref{sec5}. The sensor model described in this work considers intermittent measurements due to potential target loss caused by the limited FOV and occlusions, which is defined as
\begin{equation}
\mathbf{y}_k=\left\{ \begin{aligned}	&\mathbf{f}^s(\mathbf{z}_{k}^{t},\mathbf{z}_{k}^r)+\mathbf{w} ^s,\  &\mathbf{x}^t_k\in \mathcal{V}_k^v\\
	&\varnothing ,\qquad\quad\qquad\ \ \   &\mathbf{x}^t_k\notin \mathcal{V}_k^v\label{sensor}
\end{aligned} \right., 
\end{equation}
where $\mathbf{y}_k\in \mathbb{R}^{d_m}$ is the measurement, the measurement function 
$\mathbf{f}^s$ will be specified in \cref{sec5}, and $\mathbf{x}^t_k\in \mathbb R^2$ denotes target position. 
The sensor noise $\mathbf{w}^s$ follows a Gaussian distribution with a zero mean and covariance matrix $\mathbf{R}^s$. Here $\mathcal{V}^v_k$ is a subset of $\mathcal{V}_k$ that is not occluded by any obstacles\footnote{This work adopts a perfect sensor model for the purpose of simplicity, which assumes that a non-empty measurement is returned if and only if the target is inside the FOV and not occluded. However, it is worth noting that the proposed approach can be easily extended to imperfect sensor models that give false positive or false negative measurements.}.

\subsection{Target State Estimation}\label{EKF with im}
The robot needs accurately estimate the target state to make informed tracking behavior. To account for the system nonlinearity and possible target loss, we develop a variant of EKF for target state estimation to take intermittent measurements into account, inspired by~\cite{sinopoli2004kalman}. 

The filtering procedure can be summarized as follows.\\
$\textbf{Prediction}$.
Predict the prior PDF using target kinematics,
\begin{subequations}\label{target propagate}
\begin{align}
\hat{\mathbf{z}}^t_{k|k-1}&=\mathbf{f}^t(\hat{\mathbf{z}}^t_{k-1|k-1},\mathbf{u}^t_{k-1}),\label{KFbegin}\\
    \mathbf{P}_{k|k-1}&=\mathbf{A}^t_{k-1}\mathbf{P}_{k-1|k-1}(\mathbf{A}^t_{k-1})^T+\mathbf{R}^t,\label{KFbegin2}
\end{align}
\end{subequations}
    where $\hat{\mathbf{z}}^t_{k-1|k-1}$ and $\mathbf{P}_{k-1|k-1}$ represent the mean and covariance of the estimate of $\mathbf{z}^t_{k-1}$, and
$\mathbf{A}^t_{k-1} = \nabla_{\mathbf{z}^t}\mathbf{f}^t(\mathbf{z}^t,\mathbf{u}^t_{k-1})|_{\mathbf{z}^t=\hat{\mathbf{z}}^t_{k-1|k-1}}$ is the Jacobi matrix of the target kinematic model. The $\mathbf{u}^t_{k-1}$ can be estimated by techniques such as displacement differentiation, Linear Quadratic Gaussian Controller, etc., and is not the focus of this work. \\
$\textbf{Update}$. Use current measurements to update target PDF,
\begin{subequations}
\begin{align}
    \mathbf{K}_{k}&=\mathbf{P}_{k|k-1}\mathbf{C}_{k}^{T}(\mathbf{C}_{k}\mathbf{P}_{k|k-1}{\mathbf{C}^T_{k}}+\mathbf{R}^s)^{-1},\label{KF3}\\
\hat{\mathbf{z}}^t_{k|k}&=\hat{\mathbf{z}}^t_{k|k-1}+\mu_{k}\mathbf{K}_{k}(\mathbf{y}_{k}-\mathbf{f}^s(\hat{\mathbf{z}}_{k|k-1}^t,\mathbf{z}_{k}^r)),\label{KF4}\\
    \mathbf{P}_{k|k}&=\mathbf{P}_{k|k-1}-\mu _{k}\mathbf{K}_{k}\mathbf{C}_{k}\mathbf{P}_{k|k-1},\label{KFend}
\end{align}
\end{subequations}
    where $\mathbf{C}_{k} = \nabla_{\mathbf{z}^t}\mathbf{f}^s(\mathbf{z}^t,\mathbf{z}^r_{k})|_{\mathbf{z}^t=\hat{\mathbf{z}}^t_{k|k-1}}$ is the Jacobi matrix of measurement model, and $\mu_k$ is the \textit{detection variable} (DV) determining whether the target is detected, calculated as 
\begin{equation}
    \mu _k=\left\{ \begin{array}{c}
	1, \ \ \ \ \mathbf{y}_k\ne\varnothing\\
	0, \ \ \ \ \mathbf{y}_k=\varnothing\\
\end{array} \right.. \label{mu}
\end{equation}
\subsection{MPC Formulation for Trajectory Planning}
The trajectory planner is formulated as an MPC problem
\begin{subequations}\label{mpc problem}
\begin{align}
        \min_{\mathbf{u}^r_{k:k+N-1}}&J(\mathbf{b}^{r}_{k+1:k+N}, \mathbf{b}^{t}_{k+1:k+N})\label{origin_obj}\\
            s.t.\ \ \  & 
        \mathbf{b}^t_{k+i}=\mathbf{g}^t(\mathbf{b}^t_{k+i-1},\mathbf{b}^r_{k+i-1}),\label{target gen}\\
        &\mathbf{b}^r_{k+i}=\mathbf{g}^r(\mathbf{b}^r_{k+i-1},\mathbf{u}^r_{k+i-1})\label{robot gen},\\
        &\mathbf{b}^r_{k+i}\in \mathcal{B}^r,\mathbf{b}_{k+i}^t \in \mathcal{B}^t,\mathbf{u}^r_{k+i-1}\in \mathcal{U}\label{feas},\\
        &\mathbf{f}^o(\mathbf{b}^r_{k+i},\mathcal{O}_j)<0 ,\label{col_gen}
        \\  &j=1,2,\cdots,N_o,\ 
        i=1,\cdots,N,\nonumber
    \end{align} 
\end{subequations}
where $N$ stands for MPC planning horizon. The \textit{target belief state} $\mathbf{b}^t_k=[\hat{\mathbf{z}}_{k|k}^{t}, \mathbf{P}_{k|k}]$ encodes the probability distribution of the target state. 
Due to the existence of process noise, the robot state is stochastic in the predictive horizon. Therefore, we define the \textit{robot belief state} $\mathbf{b}^r_k=[{\hat{\mathbf{z}}^{r}_k}, \mathbf{Q}_{k}]$ to encode mean value $\hat{\mathbf{z}}^r_{k}$ and covariance matrix  $\mathbf{Q}_k$ of robot state. The sets $\mathcal{B}^r,\  \mathcal{B}^t$ and $\mathcal{U}$ are the feasible sets of robot belief state, target belief state and robot control input, respectively.
 The functions $\mathbf{g}^r$ and $\mathbf{g}^t$ denote belief prediction procedures, which will be described in detail in the next section. The objective function \cref{origin_obj} and the collision avoidance constraint \cref{col_gen} will be further elaborated in \cref{sec3,sec4}.
    
\section{Belief-Space Probability of Detection-Based State Prediction}\label{sec3}

\subsection{Probabilistic State Prediction in Predictive Horizon}

We employ two different EKF-based approaches for the state prediction of the robot and target, thus specifying \cref{target gen,robot gen}. Note that the lack of observation needs to be properly addressed, which differs from the estimation process \cref{KFbegin,KFbegin2,KF3,KF4,KFend}.

\textbf{Robot state prediction}. Like \cref{KFbegin,KFbegin2}, we use the prediction step of EKF to predict the robot state in the predictive horizon. Specifically, \cref{robot gen} is specified as
\begin{subequations}\label{robot propagate}
\begin{align}
\hat{\mathbf{z}}_{k+i}^r&=\mathbf{f}^r(\hat{\mathbf{z}}^r_{k+i-1},\mathbf{u}_{k+i-1}^r),\label{robot mean propagate}\\
\mathbf{Q}_{k+i}&=\mathbf{A}^r_{k+i-1}\mathbf{Q}_{k+i-1}(\mathbf{A}^{r}_{k+i-1})^T+\mathbf{R}^r, 
\end{align}
\end{subequations}
where $\mathbf{A}^{r}_{k+i-1}=\nabla _{\mathbf{z}^r}\mathbf{f}^r(\mathbf{z}^r,\mathbf{u}^r_{k+i-1})|_{\mathbf{z}^r=\hat{\mathbf{z}}^r_{k+i-1}}$ is the Jacobi matrix of the robot kinematic model.

\textbf{Target state prediction}. 
Within the planning horizon, the target mean is simply propagated using its kinematics with the same control input $\mathbf{u}^t_{k+i}=\mathbf{u}^t_{k-1},i=0,\cdots, N-1$, and the covariance $\mathbf{P}_{k+i|k+i-1}$ is predicted by \cref{KFbegin2}. However, target visibility is uncertain in the predictive horizon, making it difficult to predict DV and update the covariance following \cref{KF3,KFend}. 
To deal with this difficulty, we propose the concept of BPOD to denote the probability that the target is detected, defined as 
\begin{equation}
\gamma_k=Pr(\mathbf{y}_k\ne\varnothing| \mathbf{b}^r_{k},\mathbf{b}^t_{k|k-1}),\label{gamma def}
\end{equation}
where $\mathbf{b}^t_{k|k-1}=[\hat{\mathbf{z}}_{k|k-1}^t, \mathbf{P}_{k|k-1}]$. Compared to traditional PODs that are either determined by the ground truth of robot and target positions~\cite{papaioannou2019jointly} or only depend on the predictive target belief under deterministic robot states~\cite{yu2014cooperative,zhang2019multiple}, BPOD is conditioned on the belief states of both the robot and the target, which provides a more precise measurement of visibility under stochastic system states. Next, 
 we will incorporate the BPOD into EKF and constitute the stochastic counterpart of \cref{KFend} to tackle uncertain predictive visibility.

Recall that the DV is determined by the relative pose of the robot and the target, and thus can be reformulated as 
\begin{equation}
    \mu_k=Pr(\mathbf{y}_k\ne\varnothing| \mathbf{z}^r_{k},\mathbf{z}^t_{k}).\label{detection variable def}
\end{equation}
Denote $\mathbb{E}_{z|b}(\cdot)$ as the expectation operator with respect to $\mathbf{z}^t_k$ and $\mathbf{z}^r_k$ conditioned on $\mathbf{b}^t_{k|k-1}$ and $\mathbf{b}^r_k$, we can adopt the total probability rule (TPR) and derive $\gamma_k=\mathbb{E}_{z|b}(\mu_k)$. Combining \cref{detection variable def} and the EKF procedures, we can find that $\mathbf{P}_{k|k}$ in \cref{KFend} is conditioned on $\mathbf{z}^r_{k}$ and $\mathbf{z}^t_{k}$. 
Note that accurate estimates of the robot and target states are not available in the predictive horizon.
Therefore, the posterior covariance $\tilde{\mathbf{P}}_{k|k}$ in predictive horizon only depends on the predictive robot and target beliefs, and thus can be formulated as the conditional expectation of $\mathbf{P}_{k|k}$, \ie, $\tilde{\mathbf{P}}_{k|k}=\mathbb{E}_{z|b}(\mathbf{P}_{k|k})$ according to TPR. 
Following this idea, we propose a visibility-aware covariance update scheme, which is formulated as follows:
\begin{subequations}\label{BPOD updating:all}
    \begin{align}
    \tilde{\mathbf{P}}_{k|k}&=\mathbb{E}_{z|b}(\mathbf{P}_{k|k})\\
    &\approx\mathbb{E}_{z|b}(\mathbf{P}_{k|k-1}-\mu _{k}\tilde{\mathbf{K}}_{k}\tilde{\mathbf{C}}_{k}\mathbf{P}_{k|k-1})\label{BPOD updating:approxC}\\
    &=\mathbf{P}_{k|k-1}-\mathbb{E}_{z|b}(\mu_{k})\tilde{\mathbf{K}}_{k}\tilde{\mathbf{C}}_{k}\mathbf{P}_{k|k-1}\label{BPOD updating:independent}\\
    &=\mathbf{P}_{k|k-1}-\gamma_{k}\tilde{\mathbf{K}}_{k}\tilde{\mathbf{C}}_{k}\mathbf{P}_{k|k-1},\label{BPOD updating}
\end{align}
\end{subequations}
where covariance $\mathbf{P}_{k|k-1}$ and $\mathbf{P}_{k|k}$ are defined in \cref{KFbegin2,KFend}, respectively, and $\tilde{\mathbf{C}}_{k}$, $\tilde{\mathbf{K}}_k$ denote the approximate measurement Jacobi matrix and Kalman gain, respectively. 
To simplify the computation, we approximate the measurement Jacobi matrix as being equal to the one determined by the mean value of robot belief $\hat{\mathbf{z}}_k^r$, \ie, $\tilde{\mathbf{C}}_k=\nabla_{\mathbf{z}^t}\mathbf{f}^s(\mathbf{z}^t,\hat{\mathbf{z}}^r_{k})|_{\mathbf{z}^t=\hat{\mathbf{z}}^t_{k|k-1}}$, and calculate the Kalman gain $\tilde{\mathbf{K}}_k$ from \cref{KF3} by replacing $\mathbf{C}_k$ with $\tilde{\mathbf{C}}_k$,  which yields \cref{BPOD updating:approxC}. \cref{BPOD updating:independent} is derived by the independency of $\tilde{\mathbf{K}}_k$ and $\tilde{\mathbf{C}}_k$ from $\mathbf{z}^t_k$ and $\mathbf{z}^r_k$.

\begin{figure}[!t]
\centering
\includegraphics[width=\linewidth]{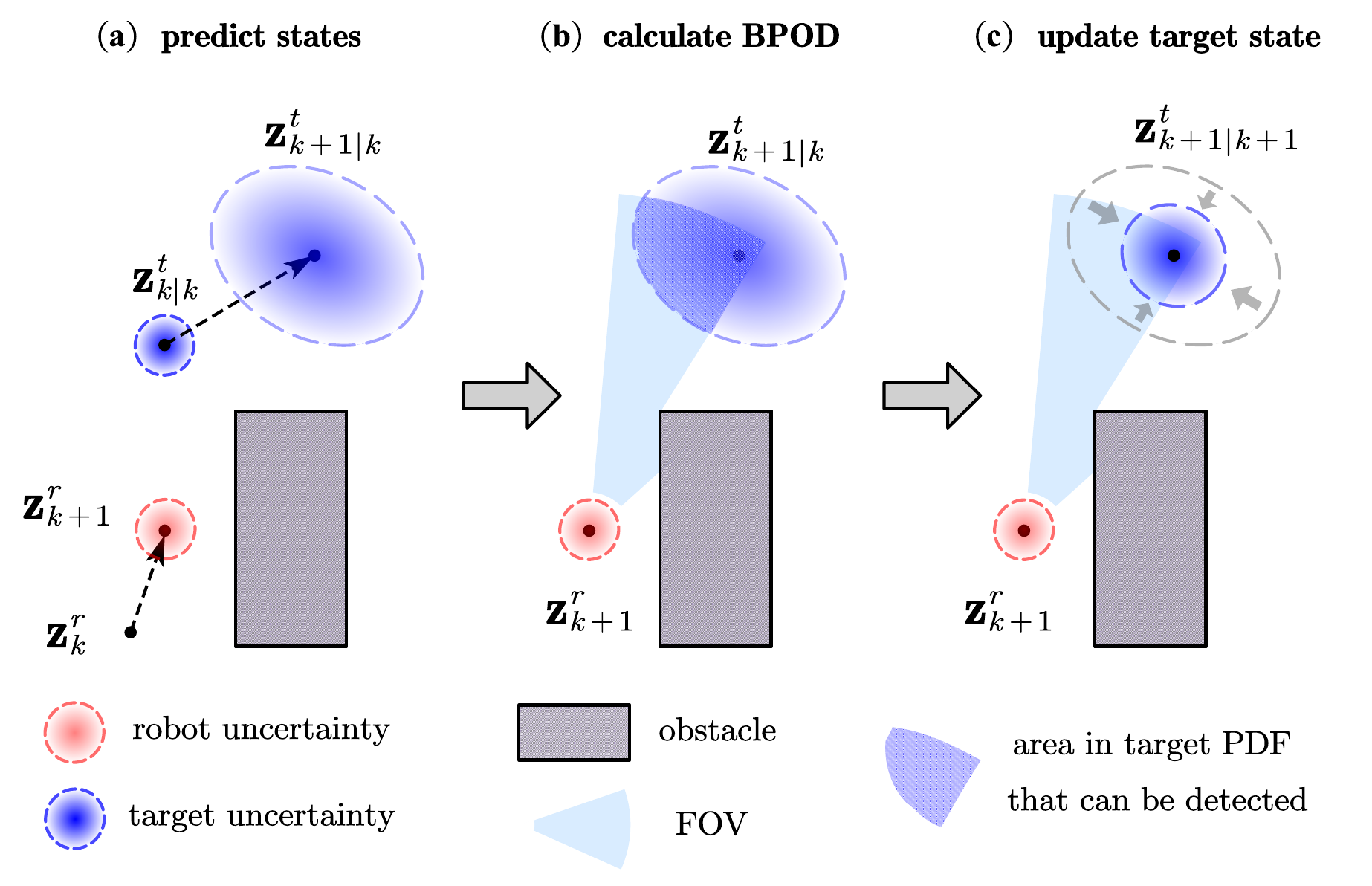}
\caption{\textbf{Illustration of state prediction process \cref{target gen,robot gen}.} (a)
Propagate robot and target prior distribution via \cref{robot propagate,target propagate}. (b) Calculate BPOD. (c) Update target covariance using \cref{BPOD updating}.}
 \label{fig:state prediction}
 \end{figure}

\cref{BPOD updating} is a probabilistic extension of \cref{KFend} that allows us to update the target covariance in \cref{target gen} using only the predicted beliefs of the robot and target. \cref{fig:state prediction} illustrate the prediction of both robot and target belief states. 

\subsection{Objective Functions}
 There are two mainstream choices of the objective function for target tracking, \ie, to minimize target uncertainty~\cite{liu2017model,zhang2019multiple}, or to maximize the predictive visibility of the target~\cite{yu2014cooperative}. We correspondingly formulate two objectives as
\begin{equation}
     J_1=\sum_{i=1}^{N} \mathcal{H}(\mathbf{b}^t_{k+i}),\ \ 
     J_2=-\sum_{i=1}^{N} \gamma_{k+i},\label{obj12}
\end{equation}
where $J_1$ is the cumulative entropy of target belief, specified as 
    $J_1 = \frac{Nd_t}{2}(\ln (2\pi)+1)+\sum_{i=1}^{N}\frac{1}{2}\ln|\tilde{\mathbf{P}}_{k+i|k+i}|$ with $d_t$ being the dimension of $\mathbf{z}^t_k$, 
and $J_2$ is the negation of cumulative BPOD. These two objectives are theoretically proved to be compatible\cite{gao2023probabilistic}, and both $J_1$ and $J_2$ are verified to perform well in keeping the target visible and decreasing estimation error in the simulation, as will be presented in \cref{sec5}.
 
\section{Signed Distance Function-Based Online Trajectory Planning}
In order to solve the MPC problem \cref{mpc problem}, it is crucial to efficiently compute BPOD and the collision constraint \cref{col_gen}. A main contribution of this work is to represent and compute the BPOD and collision risk in a unified manner that significantly reduces the computational burden of solving the MPC problem, which will be detailed in this section.
\label{sec4}
\subsection{Unified Expression of Visibility and Collision Risk}\label{collision avoidance sec}

Limited FOV and occlusion are two major factors that influence visibility. Following this idea, 
we factorize the BPOD into two types of probabilities as follows.

\textbf{Target and FOV}. We define the probability of the target being within the FOV area as:
\begin{equation}
\gamma^{tf}_k=Pr(\mathbf{x}^t_k\in \mathcal{V}_k|\mathbf{b}_k^r,\mathbf{b}_{k|k-1}^t).\label{gam_tf}
\end{equation}
\textbf{LOS and Obstacle}. Likewise, we express the probability that the target is not occluded by obstacle $i$ as:
\begin{equation}
\small
\gamma^{lo}_{k,i}=Pr(\mathcal{L}_k\bigcap\mathcal{O}_i=\varnothing|\mathbf{b}_k^r,\mathbf{b}_{k|k-1}^t), \label{gam_so}
\end{equation}
where $\mathcal{L}_k\subset\mathbb{R}^2$ is the line segment between $\mathbf{x}^t_k$ and the robot position $\mathbf{x}^r_k\in \mathbb{R}^2$. Reasonably, we can assume that all variables in $\{\gamma^{tf}_k, \gamma^{lo}_{k,1},\cdots,\gamma^{lo}_{k,N_o}\}$ are mutually independent. Then the BPOD \cref{gamma def} can be factorized as
\begin{equation}
      \gamma_k =\gamma^{tf}_{k}\prod _{i=1}^{N_o}\gamma^{lo}_{k,i}. \label{gamma comp}
\end{equation}
To avoid overly conservative trajectory while ensuring safety, we define chance-constrained~\cite{blackmore2011chance} collision avoidance conditions that bound the collision risk below a user-defined threshold $\delta^s\in \mathbb{R}$. The probability of collision between the robot and obstacle $i$ at time $k$ is defined as:
\begin{equation}
    \gamma^{ro}_{k,i}=Pr(\mathbf{x}^r_k\in \mathcal{O}_i|\mathbf{b}_k^r,\mathbf{b}_{k|k-1}^t). \label{gam_ro}
\end{equation}
\cref{col_gen} is then specified as the chance constraints, \ie,
\begin{equation}
    \gamma^{ro}_{k+i,j}<\delta^{s}_{k+i,j}\label{chance constraint}.
\end{equation}

In the next subsection, a computationally efficient algorithm is designed to calculate $\gamma^{tf}$, $\gamma^{lo}$ and $\gamma^{ro}$. For simplicity, we define the set $\Gamma_k=\{\gamma^{tf}_k, \gamma^{lo}_{k,i}, \gamma^{ro}_{k,i}, i=1,\cdots,N_o\}$.

\subsection{Approximating \text{$\Gamma_k$} with Linearized SDF} \label{gamma approximation sec}

 The SDF quantifies the distance between two shapes. Specifically, the SDF of two sets $\mathcal{A},\mathcal{B}\subset \mathbb{R}^2$ is formulated as the minimum translation distance required to separate or intersect each other, \ie,
\begin{equation*}
    sd\left( \mathcal{A} , \mathcal{B} \right) =\left\{ \begin{array}{r}
	\inf\left\{ \left\| \mathbf{v} \right\| \,|\left( \mathbf{v}+\mathcal{A} \right) \bigcap{\mathcal{B} \ne \varnothing} \right\} ,  \mathcal{A} \bigcap{\mathcal{B}}=\varnothing\\
	-\inf\left\{ \left\| \mathbf{v} \right\| \,|\left( \mathbf{v}+\mathcal{A} \right) \bigcap{\mathcal{B} =\varnothing} \right\} ,   \mathcal{A} \bigcap{\mathcal{B}}\ne \varnothing\\
\end{array} \right.\!\!\!,
\end{equation*}
where $\mathbf{v}\in \mathbb{R}^2$ is the translation vector. Using SDF, we can equivalently reformulate \cref{gam_tf,gam_so,gam_ro} as 
\begin{subequations}
    \begin{align}
    \gamma^{tf}_k &= Pr(sd(\mathbf{x}^t_k,\mathcal{V}_k)\le0|\mathbf{b}_k^r,\mathbf{b}_{k|k-1}^t),\label{sdtf}\\
    \gamma^{lo}_{k,i}&= Pr(sd(\mathcal{L}_k,\mathcal{O}_i)\ge 0|\mathbf{b}_k^r,\mathbf{b}_{k|k-1}^t),
    \label{sdlo}\\
    \gamma^{ro}_{k,i}&= Pr(sd( \mathbf{x}^r_k,\mathcal{O}_i)\le 0|\mathbf{b}_k^r,\mathbf{b}_{k|k-1}^t).
    \label{sdro}
\end{align}
\end{subequations}
Note that \cref{sdtf,sdlo,sdro} are all PoSSDFs because $\mathbf{x}^t_{k}, \mathbf{x}^r_{k}, \mathcal{L}_{k}$ and $\mathcal{V}_{k}$ are all random variables given robot and target beliefs. These probabilities can be evaluated using Monte-Carlo simulation, but at the cost of heavy computational burden. Inspired by~\cite{schulman2014motion}, we propose a computationally efficient paradigm to evaluate PoSSDF using a linearized SDF expression, as described in \cref{gamma cal}.

\begin{figure}[!t]
\centering
\includegraphics[width=\linewidth]{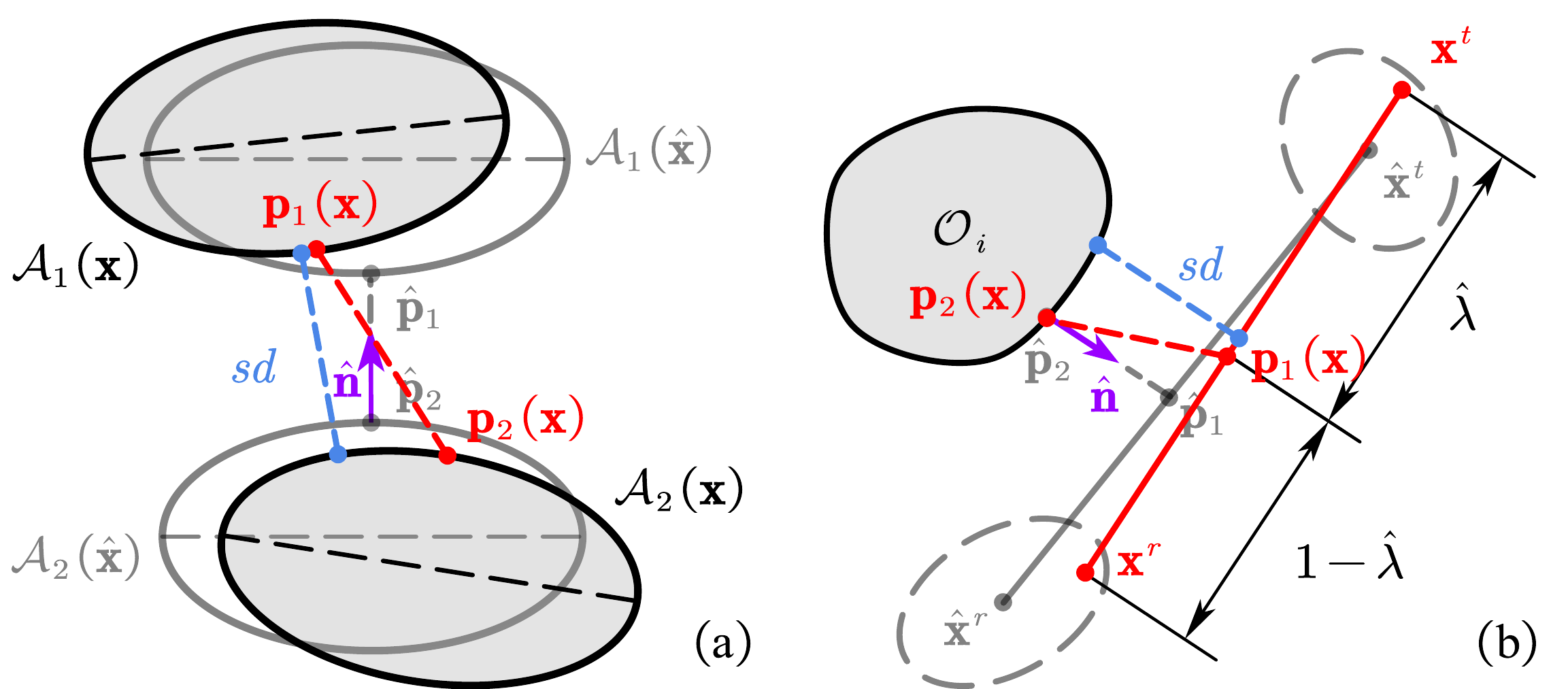}
\caption{\textbf{Illustration of SDF approximation in \cref{gamma cal}.} The blue segment represents the true SDF value of two sets $\mathcal{A}_1(\mathbf{x})$ and $\mathcal{A}_2(\mathbf{x})$, while the approximate SDF is calculated by projecting $\mathbf{p}_1(\mathbf{x})-\mathbf{p}_2(\mathbf{x})$ (the red dashed segment) to $\hat{\mathbf{n}} 
 $ (the purple vector). (a) A general case. (b) Application of \cref{gamma cal} to calculate $\gamma^{lo}$ with the dashed ellipses denoting the uncertainties of robot and target.}
\label{PoSSDF}
\end{figure}

\begin{algorithm}[t!]
\small
\fontsize{9pt}{9pt}\selectfont
\caption{Calculation of PoSSDF\label{gamma cal}}
\label{alg:alg1}
{\textsc{Input: }}two rigid bodies: $\mathcal{A}_1(\mathbf{x}), \mathcal{A}_2(\mathbf{x})$, $\mathbf{x}\sim\mathcal{N}(\hat{\mathbf{x}},\Sigma)$, \\
\hspace{0.88cm}SDF parameters 
$\hat{p}^L_{1},\hat{p}^L_{2},\hat{\mathbf{n}}$\\

{\textsc{Output: }}$p=Pr(sd(\mathcal{A}_1(\mathbf{x}),\mathcal{A}_2(\mathbf{x}))\le0)$\\

\hspace{0.5cm}$\mathbf{p}_i(\mathbf{x})\leftarrow \mathbf{R}_i(\mathbf{x})\hat{p}_i^L+\mathbf{p}^c_i(\mathbf{x}),i=1,2$\label{fixLocal}\\

\hspace{0.5cm}$sd(\mathbf{x})\leftarrow\hat{\mathbf{n}}^T(\mathbf{p}_1(\mathbf{x})-\mathbf{p}_2(\mathbf{x}))$\label{approxSDF}\\
\hspace{0.5cm}$sd_L(\mathbf{x})\leftarrow sd(\hat{\mathbf{x}})+\nabla_{\mathbf{x}}sd(\mathbf{x})|_{\mathbf{x}=\hat{\mathbf{x}}}(\mathbf{x}-\hat{\mathbf{x}})$\label{linearizeSDF}\\

\hspace{0.5cm}$p\leftarrow  \text{Calculate }  Pr(sd_L(\mathbf{x})\le0)$\label{getPossdf}
\end{algorithm}
The inputs of \cref{gamma cal} include two rigid bodies $\mathcal{A}_1,\mathcal{A}_2\subset\mathbb{R}^2$, whose poses are determined by an arbitrary random Gaussian variable $\mathbf{x}$ with dimension $d_x$. This algorithm also needs to precalculate the signed distance $\hat{d}\in\mathbb{R}$ between $\mathcal{A}_1(\hat{\mathbf{x}})$ and $\mathcal{A}_2(\hat{\mathbf{x}})$, along with the closest points from each set, $\hat{\mathbf{p}}_{1}\in \mathcal{A}_1(\hat{\mathbf{x}}),\hat{\mathbf{p}}_{2}\in\mathcal{A}_2(\hat{\mathbf{x}})$, as illustrated in \cref{PoSSDF}(a). This operation can be efficiently carried out by using Gilbert–Johnson–Keerthi (GJK) algorithm~\cite{gilbert1988fast} and Expanding Polytope Algorithm (EPA)~\cite{van2001proximity}. Then we obtain the \textit{SDF parameters} including the contact normal $\hat{\mathbf{n}}=sgn(\hat{d})\cdot(\hat{\mathbf{p}}_{1}-\hat{\mathbf{p}}_{2})/\|\hat{\mathbf{p}}_{1}-\hat{\mathbf{p}}_{2}\|$, and the local coordinates of $\hat{\mathbf{p}}_{1}$ and $\hat{\mathbf{p}}_{2}$ relative to $\mathcal{A}_1(\hat{\mathbf{x}})$ and $\mathcal{A}_2(\hat{\mathbf{x}})$ respectively, noted as $\hat{p}^L_1$ and $\hat{p}^L_{2}$.
\cref{fixLocal} provides an analytical approximation of the closest points $\mathbf{p}_1(\mathbf{x})$ and $\mathbf{p}_2(\mathbf{x})$ in the world frame. 
Here we assume that the local coordinates of $\mathbf{p}_1(\mathbf{x})$ and $\mathbf{p}_2(\mathbf{x})$, relative to $\mathcal{A}_1(\mathbf{x})$ and $\mathcal{A}_2(\mathbf{x})$ respectively, are fixed and equal to $\hat{p}^L_1$ and $\hat{p}^L_{2}$. 
The approximate closest point $\mathbf{p}_i (\mathbf{x})$ in the world frame can then be obtained by using $\mathbf{R}_i(\mathbf{x})$, the rotation matrix of $\mathcal{A}_i(\mathbf{x})$, and $\mathbf{p}_i^c(\mathbf{x})$, the origin of the local coordinate system attached to $\mathcal{A}_i(\mathbf{x})$. 
In \cref{approxSDF}, the SDF between $\mathcal{A}_1(\mathbf{x})$ and $\mathcal{A}_2(\mathbf{x})$ 
is approximated by projecting the distance between $\mathbf{p}_1(\mathbf{x})$ and $\mathbf{p}_2(\mathbf{x})$ onto the contact normal $\hat{\mathbf{n}}$. \cref{PoSSDF}(a) illustrates the SDF approximation in \cref{fixLocal,approxSDF}. In \cref{linearizeSDF,getPossdf}, we linearize the SDF around the Gaussian mean and analytically calculate the PoSSDF using the fact that the probability of a linear inequality for a Gaussian variable can be explicitly expressed as follows~\cite{prekopa2013stochastic},
\begin{equation}
   Pr(\mathbf{a}^T\cdot \mathbf{x}\le b)=\frac{1}{2}(1-\text{erf}(\frac{\mathbf{a}^T\cdot\hat{\mathbf{x}}-b}{\sqrt{2\mathbf{a}^T\Sigma\mathbf{a}}})), \label{linear pdf constr}
\end{equation}
where $\text{erf}(\cdot)$ represents the Gauss error function, and $\mathbf{a}\in\mathbb{R}^{d_{x}},b\in\mathbb{R}$ form the linear constraint of $\mathbf{x}$.

\cref{gamma cal} provides a general procedure to calculate PoSSDF. Denote $\mathbf{x}=\left[ {\mathbf{z}_{k}^{r}}^T, {\mathbf{z}_{k}^{t}}^T \right]^T$, we can make the following adjustment and apply \cref{gamma cal} to efficiently calculate $\Gamma_k$. To take the variable length of the LOS $\mathcal{L}_k(\mathbf{x})$ into account when calculating $\gamma^{lo}$, we precalculate the separative ratio $\hat{\lambda}$ of $\mathcal{L}_k(\hat{\mathbf{x}})$ before carrying out \cref{gamma cal}, which is formulated as 
 $   \hat{\lambda}=\|\hat{\mathbf{p}}_1-\hat{\mathbf{x}}^t_k\|/\|\hat{\mathbf{x}}^r_k-\hat{\mathbf{x}}^t_k\|,$
 where $\hat{\mathbf{p}}_1$ is the nearest point on $\mathcal{L}_k(\hat{\mathbf{x}})$. We then replace the SDF parameter $\hat{p}_1^L$ with $\hat{\lambda}$ and reformulate the approximated nearest point (\cref{fixLocal}) on $\mathcal{L}_k(\mathbf{x})$ in the world frame as
\begin{equation}
\mathbf{p}_1(\mathbf{x})=\hat{\lambda}\mathbf{x}^r_k(\mathbf{x})+(1-\hat{\lambda})\mathbf{x}^t_k(\mathbf{x}).\label{lambda}\\
\end{equation}
This adjustment for calculating $\gamma^{lo}$ is illustrated in \cref{PoSSDF}(b). 

Note that \cref{chance constraint} is not violated even though $\gamma^{ro}$ is approximated with \cref{gamma cal}. This is because in \cref{gamma cal}, the obstacle is expanded to a half space with the closest point on its boundary and $\hat{\mathbf{n}}$ being its normal vector, which overestimates $\gamma^{ro}$ thus ensuring \cref{chance constraint} to be strictly enforced.

\subsection{Sequential Convex Optimization}\label{subsec:scp}
After being specified in \cref{collision avoidance sec,gamma approximation sec}, problem \cref{mpc problem} can be solved using the SCP algorithm, where the nonlinear constraints are converted into $l_1$ penalty functions,
\begin{equation}
J_m=J+\eta(\sum_i|g^n_i|^++\sum_i|h^n_i|).\label{mod J}
\end{equation}
Here $J$ denotes the objective function \cref{origin_obj}, and $g_i^n$, $h_i^n$ represent the nonlinear inequality and equality constraints, respectively. Here $\eta$ is the penalty coefficient and $|x|^+=\max(x,0)$.
The SCP consists of two loops: 
The outer loop progressively increases $\eta$ to drive the nonlinear constraint violation to zero, while in the inner loop, the trust region method is implemented to minimize the new objective function ${J}_m$. Interested readers can refer to~\cite{schulman2014motion} for details. 

Directly adopting this framework turns out to be slow due to the frequent calls of GJK algorithms and EPA when calculating the gradient of \cref{mod J}. To overcome this limitation, we use a fixed SDF parameter set that encodes all the SDF parameters in the MPC horizon to calculate $\Gamma_{k+1:k+N}$ in the gradient, and update the SDF parameter set after obtaining a new solution. This two-stage strategy significantly speeds up the planner with minor accuracy loss, and is further described in the trajectory planning framework presented in \cref{alg:alg2}.

\begin{algorithm}[!t]
\small
\fontsize{9pt}{9pt}\selectfont
\caption{Trajectory Planning in SCP Framework with Two-Stage Strategy for Efficient SDF Calculation}
\label{alg:alg2}
\textsc{Parameters: }\\
\hspace{0.5cm}$\eta:\text{initial penalty coefficient}$\\
\hspace{0.5cm}$d:\text{initial trust region size}$\\
\hspace{0.5cm}$\beta:\text{increase ratio of the penalty coefficient}$\\
\hspace{0.5cm}$\tau_c,\tau_p,\tau_f:\text{convergence tolerances}$\\
{\textsc{Output: }}$x^\ast=\mathbf{u}^r_{k:k+N-1}:\text{the optimal solution} $\\
\nonl\hrulefill  \\
$x_0,\mathcal{C}_0\leftarrow$ Initialize the solution and SDF parameter set\label{alg2:init}\\
\While{TRUE}{
    \While{TRUE}{
     $\nabla J_x \leftarrow$ Get the gradient of $J_m$ on $x_0$ using $\mathcal{C}_0$\label{alg2:getGrad}\\
     $\tilde{J}(x,\eta,\mathcal{C}_0)\leftarrow J_m(x_0,\eta,\mathcal{C}_0)+\nabla J_x\cdot(x-x_0)$\label{alg2:linearize}\\
     $ x^\ast\leftarrow \textbf{argmin}_x \tilde{J}(x,\eta,\mathcal{C}_0)$ subject to trust region, linear constraints, and semidefinite constraints of covariance matrices. \\
    $ \mathcal{C}^{\ast}\leftarrow$ Update SDF parameter set from $x^\ast$\label{alg2:updateC}\\
    $d\leftarrow$ Update trust region size by improvement ratio 
    $\frac{J_m(x_0,\eta,\mathcal{C}_0)-J_m(x^\ast,\eta,\mathcal{C}^{\ast})}{J_m(x_0,\eta,\mathcal{C}_0)-\tilde{J}(x^\ast,\eta,\mathcal{C}_0)}$\label{alg2:update d}\\
    $\textbf{If } \|x^\ast-x_0\|\le\tau_c\textbf{ or }|\nabla J_x\cdot(x^{\ast}-x_0)|\le\tau_f\ \textbf{break}$\label{alg2:breakInner}\\
    $x_0\leftarrow x^\ast$, $\mathcal{C}_0\leftarrow \mathcal{C}^{\ast}$ If trust region is expanded
    }
$\textbf{If} \sum_{i}|g^n_i(x^\ast,\mathcal{C}^{\ast})|^++\sum_{j}|h^n_j(x^\ast,\mathcal{C}^{\ast})|\le\tau_p\   \textbf{break}$\label{alg2:breakOuter}\\
$\eta\leftarrow \beta\cdot\eta\label{alg2:increaseIta}$
}
\label{alg1}
\end{algorithm}
The variable $x_0$ is initialized with zero control inputs at the first step, and is extrapolated from its value at the previous step for all subsequent steps. This is followed by precalculating an SDF parameter set $\mathcal{C}_0$ at the mean value of the initial robot and target beliefs, which is propagated by $x_0$ (\cref{alg2:init}). Using a fixed $\mathcal{C}_0$, we can fastly obtain the gradient (\cref{alg2:getGrad}) and linearize the objective function (\cref{alg2:linearize}). After solving the linearized problem, the SDF parameter set is updated (\cref{alg2:updateC}) and utilized to update the trust region size (\cref{alg2:update d}). The inner loop ends when the improvement is small (\cref{alg2:breakInner}). The outer loop checks the terminal condition for the solution $x^\ast$ (\cref{alg2:breakOuter}) and increases the penalty coefficient $\eta$ (\cref{alg2:increaseIta}).
\setstretch{0.99}
\section{Simulations}\label{sec5}
The proposed method is validated through multiple simulations in MATLAB using a desktop (12th Intel(R) i7 CPU@2.10GHz), and the MOSEK solver is adopted to optimize the trajectory according to the SCP routine. The robot takes a unicycle model, where the robot state $\mathbf{z}^r_k=[{\mathbf{x}^r_k}^T, \theta_k^r, v_k^r]^T\in \mathbb{R}^4$
contains position ${\mathbf{x}^r_k}\in \mathbb{R}^2$, orientation $\theta_k^r\in (-\pi,\pi]$, and velocity $v_k^r\in \mathbb{R}_+$, and the robot control input $\mathbf{u}^r_k=[\omega_k^r,a_k^r]^T\in \mathbb{R}^2$ is composed of 
angular velocity $\omega_k^r\in \mathbb{R}$ and acceleration $a_k^r\in\mathbb{R}$. The robot dynamics follow the following motion model:
\begin{equation}
\mathbf{f}^r(\mathbf{z}^r_k,\mathbf{u}^r_k)=\mathbf{z}^r_k+[v_k^r\cos\theta_k^r,\, v_k^r\sin\theta_k^r,\, \omega_k^r,\, a_k^r]^T \cdot \Delta t,
\end{equation}
with $\Delta t=0.5s$ representing our sampling interval. We set the range of robot acceleration ($m/s^2$) as $-4\le a_k^r\le 2$, angular velocity ($rad/s$) as $-\pi/3 \le \omega_k^r\le\pi/3$ and speed limit as $4m/s$. The motion noise is set as $\mathbf{R}^r=10^{-3}\cdot diag(4,4,0.4,0.4) $. The robot's sensor FOV is modeled as an annular sector, with minimal detection distance $r_1=2m$, maximal distance $r_2=10m$ and the sensing angle $\psi^s$ is $2\pi/3$. We shrink the FOV to its convex subset when calculating \cref{gam_tf} to fit the inputs of GJK algorithm and EPA. The predictive horizon is set as $N=4$. A $60m\times 50m$ map with cluttered polygon obstacles is designed in our simulation tests (\cref{linear tracking}(a)). The proposed method only considers obstacles near the LOS, which we note as ``valid obstacles''. The initial state of the robot is designated as 
 $[32,7,\frac{3}{4}\pi,0]^T$. 

 To quantitatively evaluate the performance of our method, several metrics are computed for a tracking simulation with a total step $T$ and target trajectory $\{\tilde{\mathbf{x}}^t_k,k=1,\cdots,T\}$, including computing time $t_{cal}$, loss rate $r_{los}$ that denotes the percentage of time the robot cannot see the target, and the estimation error $e_{est}$ that is defined as the mean absolute error (MAE) of the target's estimated position, \ie,
      $e_{est}=\frac{1}{T}\sum_{k=1}^T\|\hat{\mathbf{x}}^t_k-\tilde{\mathbf{x}}^t_k\|. $
Besides, we claim a \textit{tracking failure} if the robot collides with an obstacle or loses sight of the target in 15 consecutive steps, and we define the success rate $r_{suc}$ over multiple tracking experiments as the proportion that the robot finishes the tracking task without failure. 
\subsection{Performance Analysis in a Challenging Scenario}\label{simu 1}
\begin{figure}[!t]
\centering
\includegraphics[width=\linewidth]{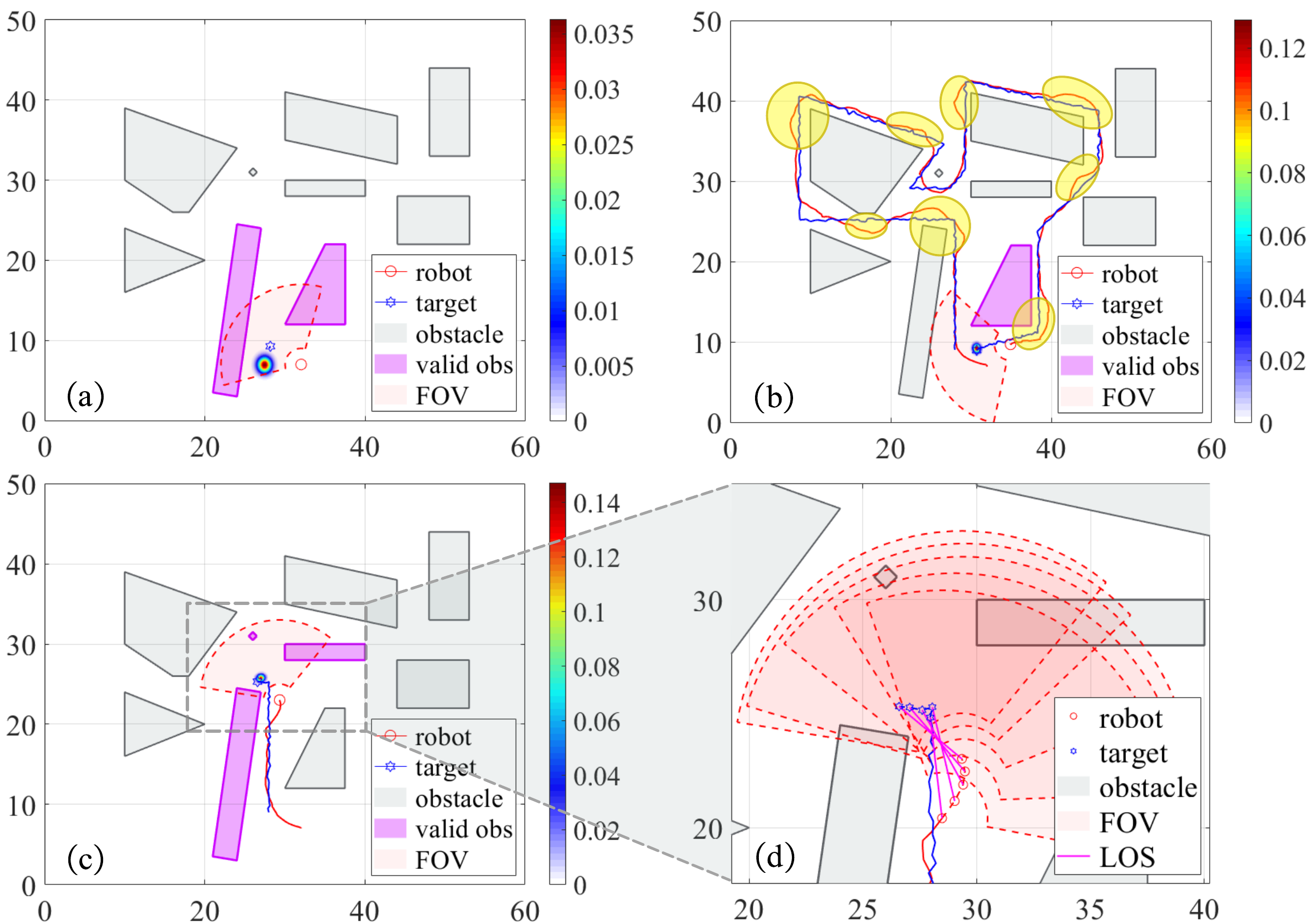}
\caption{\textbf{Visibility-aware tracking of the target with linear motion model.} The colorful background represents the target's PDF, and the color bar shows the colormap of probability. The ``valid obs'' denotes the valid obstacles. (a) Initial simulation scenario. (b) Entire tracking trajectory. (c) and (d) highlight a sharp turn of the target and display the robot's trajectory with high visibility. }
\label{linear tracking}
\end{figure}

We design a target trajectory with several sharp turns near obstacles to evaluate the performance of the tracker under challenging scenarios, as shown in \cref{linear tracking}(b). We adopt a single integrator to describe the target motion model, which is formulated as 
$\mathbf{z}^t_{k+1}=\mathbf{z}^t_k+\mathbf{u}^t_k\cdot \Delta t$, 
where the target state consists of its position, \ie, $\mathbf{z}^t_k=\mathbf{x}^t_k$, and target control $\mathbf{u}^t_k\in\mathbb{R}^2$ is known to the robot. 
Covariance $\mathbf{R}^t$ is set as $0.01\mathbf{I}_2$, where $\mathbf{I}_2$ denotes the $2\times 2$ identity matrix. A range-bearing sensor model is adopted to acquire the distance and bearing angle of the target relative to the robot, which is written as:
\begin{equation}
\mathbf{f}^s(\mathbf{z}_{k}^{t},\mathbf{z}_{k}^r)=[\|\mathbf{x}_{k}^t-\mathbf{x}_{k}^r\|, \angle (\mathbf{x}_{k}^t-\mathbf{x}_{k}^r)-\theta^r_k]^T.\label{rb sensor}
\end{equation}
The covariance $\mathbf{R}^s$ is set as $diag(0.3,0.05)
$. The target position is initialized at $[28,9]^T$. We use cumulative entropy $J_1$ in \cref{obj12} as our objective function.

\begin{figure}[!t]
\centering
\includegraphics[width=\linewidth]{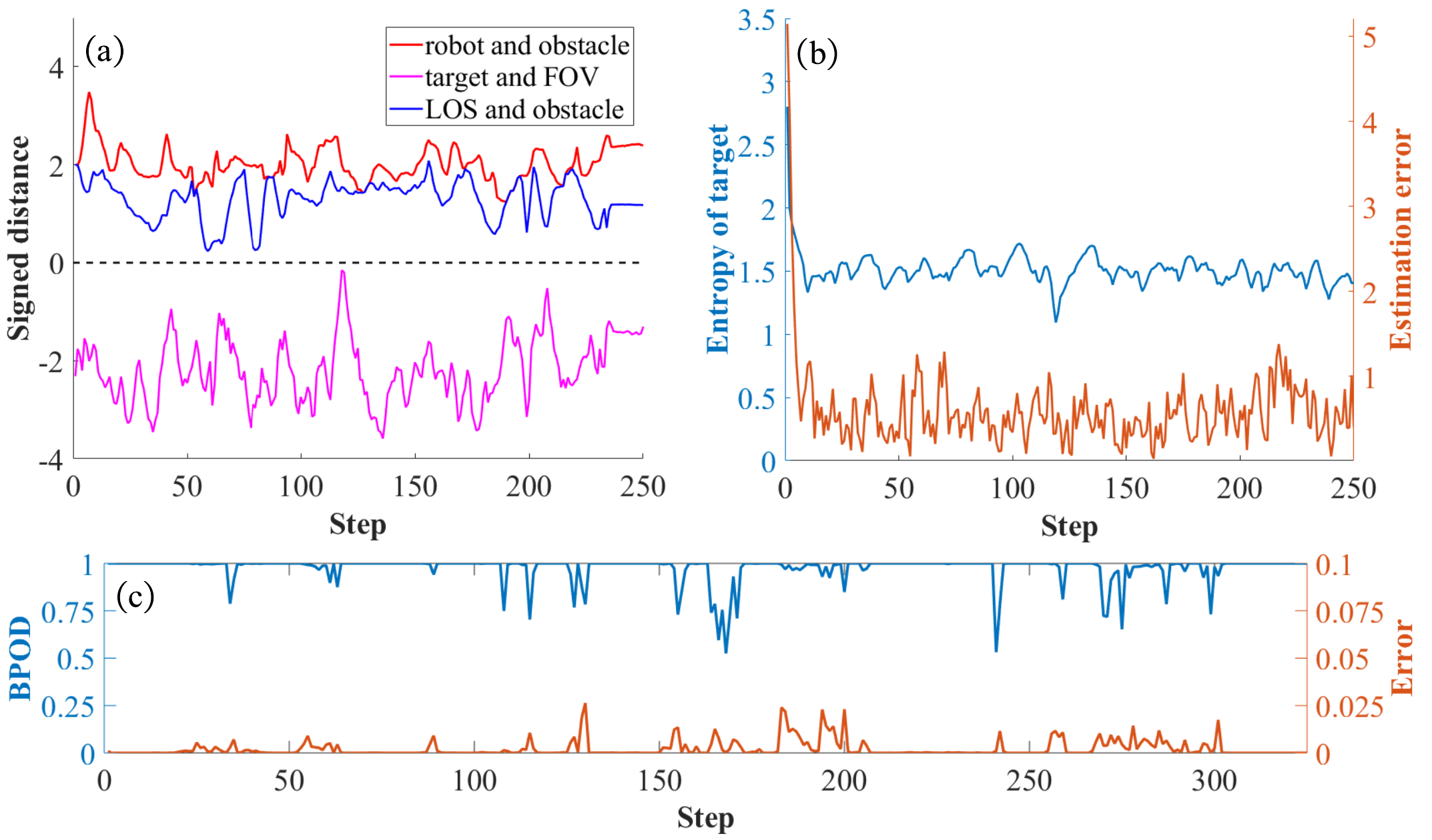}
\caption{\textbf{Trajectory performance.} (a) History of signed distances. (b)  Target entropy and estimation error. (c) The BPOD value and the approximation error of the BPOD compared to the ground truth.}
\label{lineardata}
\end{figure}

\cref{linear tracking} shows the tracking process. The target uncertainty is initialized to be very large 
(\cref{linear tracking}(a)). 
As the tracking progresses, the robot plans a visibility-aware trajectory to track the target (\cref{linear tracking}(b)). Specifically, when the target takes a sharp turn near an obstacle,  the robot moves away from the obstacle in advance to reduce the likelihood of target loss (\cref{linear tracking}(c) and (d)). Note that this roundabout path frequently appears in the tracking process (highlighted by yellow ellipses in  \cref{linear tracking}(b)) and is a characteristic of visibility-aware trajectories.

\cref{lineardata} displays the performance data.
The robot can maintain the visibility of the target throughout the simulation without any target loss. To see this,  \cref{lineardata}(a) shows the purple curve that indicates the target is inside our convexified FOV most of the time, and the blue curve that indicates LOS and obstacles never intersect. Due to uninterrupted measurements, target entropy rapidly converges and the estimation error is kept small (\cref{lineardata}(b)). 
We also record the BPOD value and evaluate the accuracy of the BPOD approximation against the Monte-Carlo simulation as the ground truth, as shown in \cref{lineardata}(c). The subfigure shows that the BPOD is kept near $1$ most of the time, and simulation results also demonstrate that we can calculate one BPOD within $1ms$ with an MAE of $2.5\times10^{-3}$.

The experiment is conducted 30 times for quantitative performance evaluation. The robot completes the tracking task in 28 cases without failure, and the results show that the robot can track the target with low mean estimation error $\bar{e}_{est}=0.308m$, low mean loss rate $\bar{r}_{los}=4.1$\% with the mean computing time $\bar{t}_{cal}= 0.176s/step$.
\subsection{Evaluation with Stochastic Target Trajectories}\label{case2}

 We benchmark our method against a representative visibility-aware trajectory planning method based on \cite{wang2021visibility}. To ensure a fair comparison, both methods utilize an MPC framework with identical robot kinematics constraints and SCP solver. The main distinction lies in that we adapt the deterministic visibility cost and the collision cost in \cite{wang2021visibility} to the baseline MPC framework as the objective function. 

The target takes a unicycle model\cite{lavalle2006planning} with the target state represented as $\mathbf{z}^t_k=[{\mathbf{x}^t_k}^T,\theta^t_k]^T$, where $\theta^t_k$ stands for target orientation. In addition, the target controls $\mathbf{u}^t_k=[v^t_k, \omega^t_k]^T$ that encode the speed $v^t_k\in \mathbb{R}$ and angular velocity $\omega^t_k\in\mathbb{R}$ remain unknown to the robot. To estimate and predict target belief according to the EKF fashion, the robot estimates $\mathbf{u}^t_k$ by differentiating target displacement and rotation. We choose the cumulated BPOD $J_2$ in \cref{obj12} as the objective function. Besides, we adopt a camera sensor model to detect target distance, bearing angle and orientation:
\begin{equation}
\mathbf{f}^s(\mathbf{z}_{k}^{t},\mathbf{z}_{k}^r)=[\|\mathbf{x}_{k}^t-\mathbf{x}_{k}^r\|, \angle (\mathbf{x}_{k}^t-\mathbf{x}_{k}^r)-\theta^r_k,  \theta^t_k-\theta^r_k]^T.\label{camera sensor}
\end{equation}

We test both our algorithm and the baseline algorithm under different scales of measurement noise. Specifically, the measurement noise is set as $10^{-2}\beta\cdot diag(1, 0.5, 1)$, with $\beta\in\{0.5,2,5\}$ corresponding to different uncertainty levels. For each level, 50 randomly generated target trajectories are tested to compare the tracking performance of the two planners. The target speed limit is set as $3m/s$ and each trajectory has $400$ time steps. Target motion noise is set as $0.5\mathbf{I}_3$ to simulate stochastic target movements. 

\begin{figure}[!t]
\centering
\includegraphics[width=\linewidth]{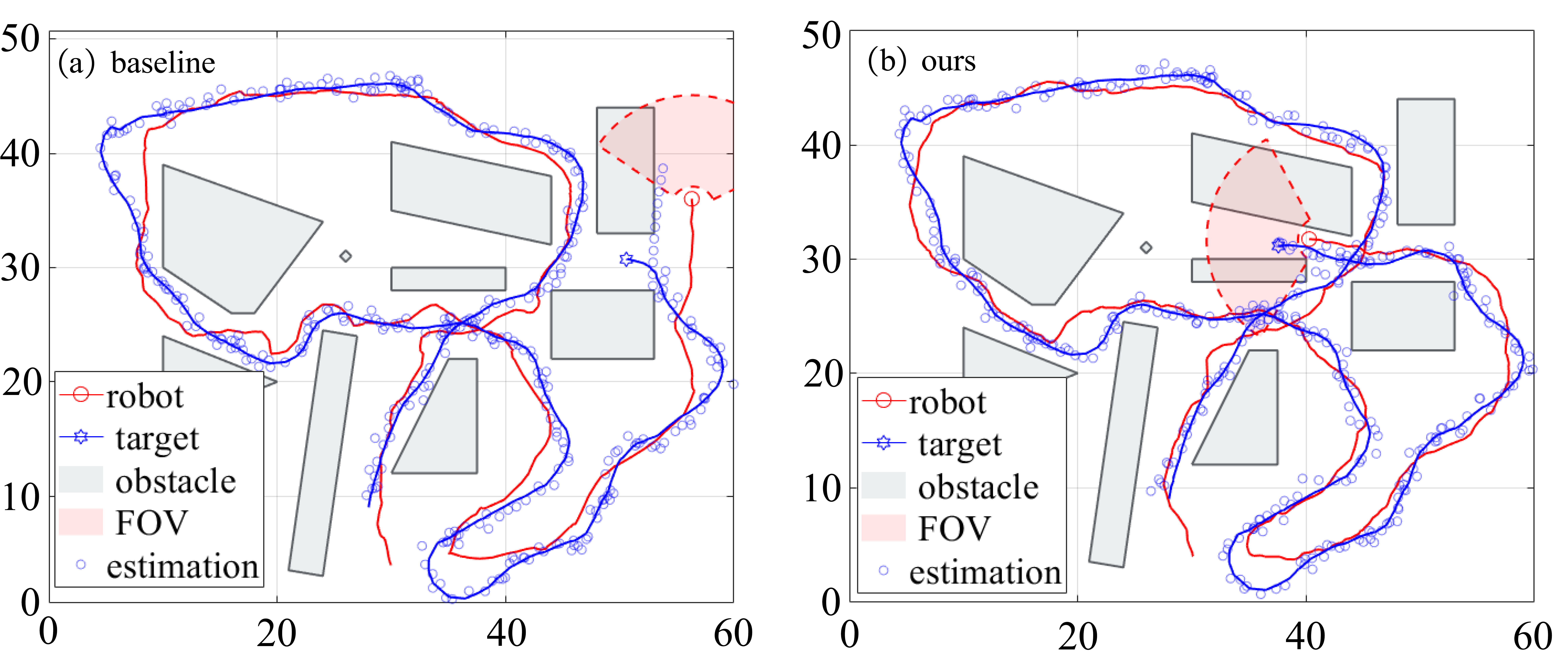}
\caption{\textbf{Trajectory comparisons between baseline method (a) and ours (b).} In the face of stochastic target trajectories and high system noise ($\beta=5$), the baseline method generates a more tortuous trajectory and finally loses the target due to inaccurate estimations. In contrast, our method generates a smoother trajectory and succeeds in completing the tracking task.}
\label{nonlinear tracking}
\end{figure}

\begin{figure}[!t]
\centering
\includegraphics[width=\linewidth]{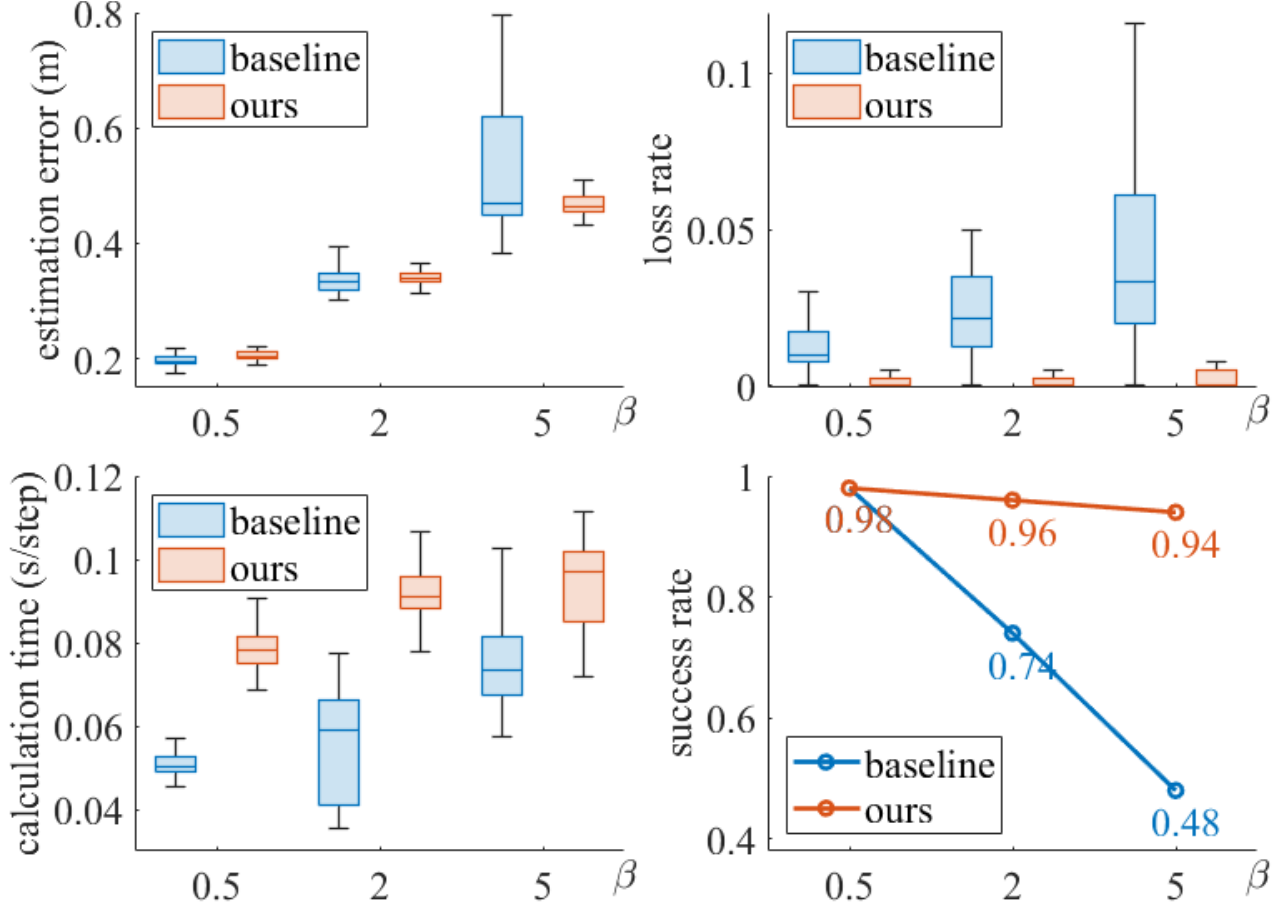}
\caption{\textbf{Box graphs of the performance metrics under different uncertainty levels $\beta$.} The estimation error, loss time and calculation time are all averaged over 50 tracking experiments.}
\label{comparison}
\end{figure}

A pair of comparative tracking experiments is presented in \cref{nonlinear tracking} to highlight the negative impact of system uncertainty on a deterministic trajectory planner and to showcase the robustness of our approach. Simulation metrics comparisons are depicted in \cref{comparison}. The results show that in low-noise scenarios, both planners provide precise estimations for target states and effectively keep the target visible. However, in the face of higher system uncertainty, the proposed planner consistently maintains low estimation error, high visible rate, and high success rates when tracking a target, while the performance of the baseline deteriorates significantly. It is important to note that although our method requires additional time for trajectory planning due to extra computational cost for considering the system uncertainty, the overall duration remains below 0.1 seconds, which is sufficient for real-time planning. The overall metrics show that our method achieves a better balance between efficiency and effectiveness in the face of high uncertainty. 

\section{Real-World Experiments}\label{sec6}
The proposed approach has been tested by using a Wheeltec ground robot equipped with an ORBBEC Femto W camera to keep track of a moving Turtlebot3, on which three interlinked Apriltags are affixed.
The speed limit of the tracker and the target are $0.4m/s$ and $0.31m/s$, respectively. The onboard processor (NVIDIA Jetson TX1) on the Wheeltec robot calculates the distance, bearing angle, and orientation of the detected Apriltag from the camera images and transmits the messages through ROS to a laptop (11th Intel(R) i7 CPU@2.30GHz) that performs the planning algorithm. The parameters of the sensor model are calibrated as $r_1=0.3m$, $r_2=1.5m$, $\theta=100^\circ$, and $\mathbf{R}^s = diag(0.069,8.3\times 10^{-4},0.0055)$. 
The covariance of robot motion noise is set as $\mathbf{R}^r=diag(0.08\mathbf{I}_2,0.055\mathbf{I}_2)$ to handle model discretization error and mechanical error. Other parameters are kept the same as \cref{case2} except for the distance parameter that has been scaled down by a factor of 10. The ground truth of the robot and target poses are obtained by a Vicon motion-capture system. 

\begin{figure}[!t]
\centering
\includegraphics[width=\linewidth]{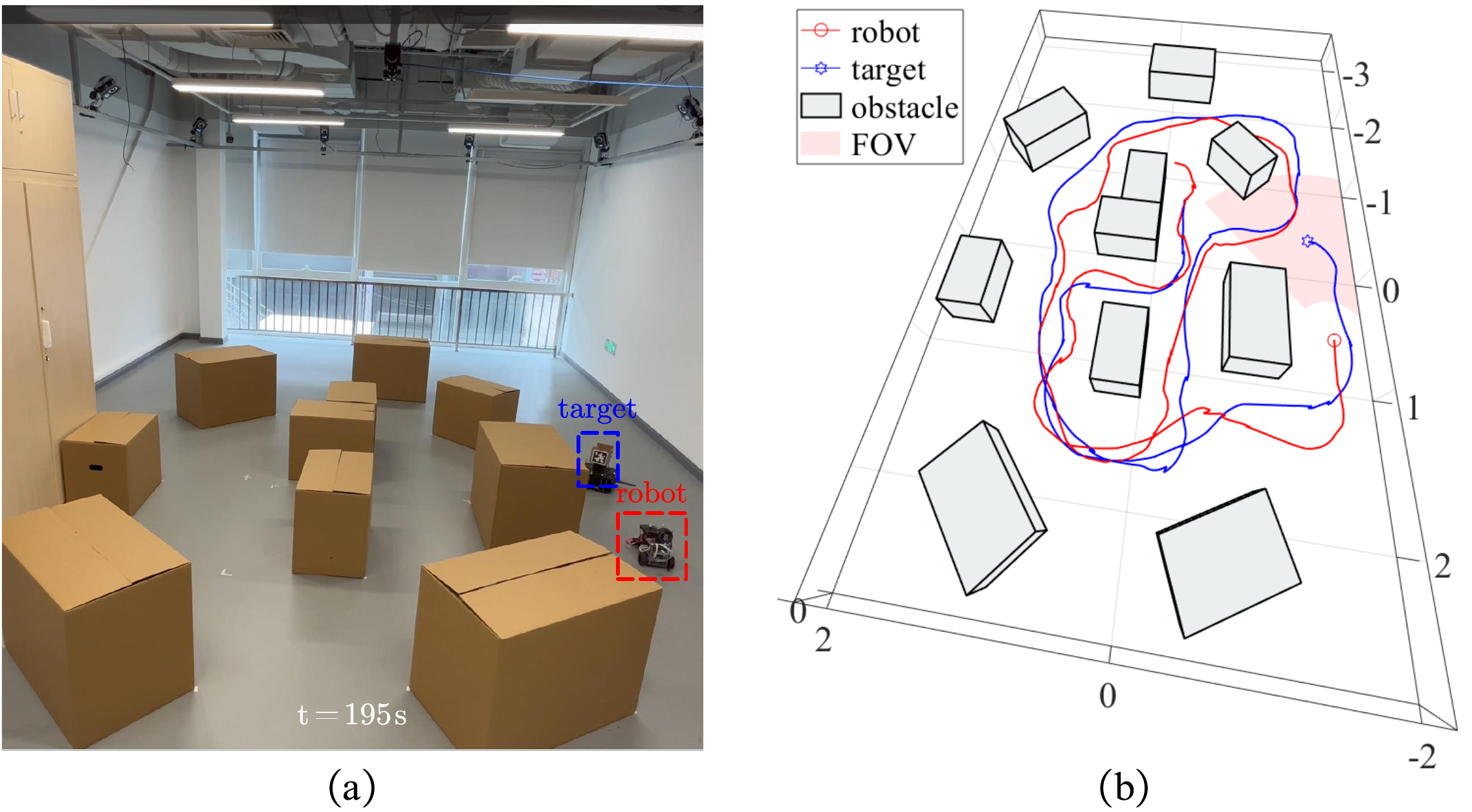}
\caption{\textbf{Real-world target tracking.} (a) Map configuration and one tracking frame. (b) Simulation scene corresponding to (a) and the trajectories of the target and the robot.}
\label{real-world}
\end{figure}

We conduct three tracking experiments by remotely controlling the target to randomly traverse an indoor map with cluttered obstacles (\cref{real-world}(a)), and the duration of all three experiments exceed 2 minutes. For each experiment, we record the total planning steps $T$, estimation error $e_{est}$, calculation time $t_{cal}$ and minimum distance to obstacles $d_{min}$, as shown in \cref{tab:real-world}. Both \cref{real-world}(b) and \cref{tab:real-world} show that our planner can generate safe trajectories for the robot to track a moving target while reducing target uncertainty in real-world scenarios. 

\begin{table}[!t]
\small
\caption{Real-World Experiment Results \label{tab:real-world}}
\centering
\begin{tabular}{|c|c|c|c|c|}
\hline
& $ T$  &  $e_{est}$(m)  & 
 $t_{cal}$(s/step)  &  $d_{min}$(m)  \\
\hline
1&538&0.072&0.087&0.247\\ 
\hline
2&497&0.078&0.089&0.210\\  
\hline
3&471&0.071&0.090&0.232 \\ 
\hline
\end{tabular}
\end{table}

\section{Conclusion}\label{sec7}
We propose a target-tracking approach that systematically accounts for the limited FOV, obstacle occlusion, and state uncertainty. In particular, the concept of BPOD is proposed and incorporated into the EKF framework to predict target uncertainty in systems subject to measurement noise and imperfect motion models. We subsequently develop an SDF-based method to efficiently calculate the BPOD and collision risk to solve the trajectory optimization problem in real time. Both simulations and real-world experiments validate the effectiveness and efficiency of our approach.

Future work includes mainly two aspects. First, we will investigate the properties and applications of BPOD in non-Gaussian belief space planning. Second, we will extend the proposed method to unknown and dynamic environments.

\textbf{Acknowledgement:}
We thank Dr. Meng Wang at BIGAI for his help with photography.
{
\setstretch{0.97}
\small
\bibliography{main}
\bibliographystyle{ieeetr}
}
\end{document}